# On Enhancing Genetic Algorithms Using New Crossovers

Ahmad B. A. Hassanat *, Esra'a Alkafaween

IT Department, Mutah University, Mutah – Karak, Jordan, 61710
*Corresponding Author: Ahmad.hassanat@gmail.com

**Abstract-** This paper investigates the use of more than one crossover operator to enhance the performance of genetic algorithms. Novel crossover operators are proposed such as the Collision crossover, which is based on the physical rules of elastic collision, in addition to proposing two selection strategies for the crossover operators, one of which is based on selecting the best crossover operator and the other randomly selects any operator.
Several experiments on some Travelling Salesman Problems (TSP) have been conducted to evaluate the proposed methods, which are compared to the well-known Modified crossover operator and partially mapped Crossover (PMX) crossover. The results show the importance of some of the proposed methods, such as the collision crossover, in addition to the significant enhancement of the genetic algorithms performance, particularly when using more than one crossover operator.

**Biographical notes**: Ahmad B. A. Hassanat was born and grew up in Jordan, received his Ph.D. in Computer Science from the University of Buckingham at Buckingham, UK in 2010, and B.S. and M.S. degrees in Computer Science from Mutah University/Jordan and Al al-Bayt University/Jordan in 1995 and 2004, respectively. He has been a faculty member of Information Technology department at Mutah University since 2010. His main interests include computer vision, Machine learning and pattern recognition.

Esra'a Alkafaween received her Master degree in Computer science from Mutah University, Jordan, in 2015. She is currently a Researcher and laboratory supervisor at Mutah University, and pursuing her research in the Department of Computer Science in the field of Artificial Intelligence.

## 1. INTRODUCTION

Genetic algorithms (GAs) are powerful heuristic random search techniques that mimic the theory of evolution and natural selection. The basic principles of the genetic algorithm were investigated by John Holland in the 1970's (Holland, 1975) in the University of Michigan. The genetic algorithm has proved its strength and durability in solving many problems, and thus it is considered as an optimization tool for many researchers (Goldberg, 1989), (Mahmoudi & Mahmoudi, 2014), (Kotenko & Saenko, 2015) and (Tsang & Au, 1996). This explains the increase and expansion of its popularity among researchers in many areas such as image processing (Paulinas & Ušinskas, 2015) and (Amsaveni & Vanathi, 2015), speech recognition (Benkhellat & Belmehdi, 2012) (Gupta & Wadhwa, 2014), software engineering (Srivastava & Kim, 2009), computer networks (Nakamura, 1997), robotics (Ayala & dos Santos Coelho, 2012), in addition to other applications such as in (Tian, et al., 2016), (Chen, et al., 2015) and (Kantha, et al., 2016).

The genetic algorithm was inspired by the Darwinian theory of "survival of the fittest" (Zhong, et al., 2005), (Mustafa, 2003) & (Eiben & Smith, 2003), by producing new chromosomes (individuals) through recombination (crossover) and mutation operations, i.e. the fittest individual is more likely to remain and mate. Therefore, the inhabitants of the next generation will be stronger, because they are produced from strong individuals, i.e. the solution evolves from one generation to another.

The GA begins with a number of random solutions (initial population), these solutions (individuals) are then encoded according to the current problem, and the quality of each individual is evaluated through a fitness function (Bäck & Schwefel, 1993). The GA depends mainly on three operators:

- Selection (competition): The process of choosing the "best" parents in the community for mating, "best" being defined based on the current problem.
- Crossover operator: Takes two parents (chromosomes), to create a new offspring by switching segments of the parent genes; it is more likely that the new offspring would contain good parts of their parents, and consequently perform better as compared to their ancestors.
- Mutation operator: Takes a single chromosome, and alters some of its genes to create a new chromosome, hopefully better than its parent.

The better choice of these processes and the better selection of their parameters (such as the crossover and mutation rates) enhance the performance of the GA (Eiben, et al., 2007). Many researchers have shown the effect of the two operators: crossover and mutation on the success of the GA, and where success lies in both, whether crossover is used alone or mutation alone or both together, as in (Spears, 1992) and (Deb & Agrawal, 1999).

Crossover operators have a role in the balance between exploitation and exploration, which will allow the extraction of characteristics from both parents, and hopefully that the resulting offspring possess good characteristics from the parents (Gallard & Esquivel, 2001).

Over the years, many types of crossover have been developed, and comparisons have been made between these types. These began with the one-point crossover, then evolved into several techniques to accommodate a number of situations, including Uniform crossover (Syswerda, 1989), Multi-point crossover (Spears & De Jong, 1990), Heuristic crossover (Grefenstette, et al., 1985), Ring crossover (Kaya & Uyar, 2011), and for the order-based problem, the Partially Matched crossover (PMX) (Goldberg & Lingle, 1985), Cycle crossover (CX) (Oliver, et al., 1987), Order crossover (OX) (Goldberg, 1989) and some other types.

Our selection of the type of crossover depends mainly on the type of encoding used. This can sometimes be very complicated but normally improves the performance of the genetic algorithm (Man, et al., 1996). However, the question of whether this kind of crossover is better than the others remains open. In this regard we cannot draw a significant conclusion about which is better, because most of the comparisons between the different types were conducted on a small group of test problems and more trial and error was needed. In order to overcome this problem, several researchers have developed new types of GA that use more than one crossover operator at the same time (Dong & Wu, 2009) (Hilding & Ward, 2005), another study of Spears in which he stated: "But much of the performance stems from simply having two crossover operators at the GA's disposal. This raises the intriguing notion that it may often be beneficial for an EA to have a larger set of search operators than is customarily used." (Spears, 1995). Therefore, our paper contributes to emphasize these conclusions.

In the recent past, a lot of effort has been put into the development of new crossover operators to improve the performance of GA (Deep & Thakur, 2007).This paper aims to provide new crossover operators, and investigates the effect of using more than one crossover operator on the performance of the GA. The contribution of this paper is two-fold: (1) proposals of new crossover operators to find better solutions for TSP, and (2) investigations into the effect of using more than one of these crossovers on the performance of the GA.

The rest of this paper presents some of the related previous work and the proposed methods, in addition to discussing the experimental results, which were designed to evaluate the proposed methods. The conclusions and the future work are presented at the end.

## 2. Related work

Vekaria et al. suggested a crossover called selective crossover that mimics Dawkins' theory of natural evolution, and in storing the knowledge of previous generations it has been used as an extra real-value vector. This crossover method was compared with two types of crossovers: two-point and uniform crossover, and the results showed that this outperformed some of the other methods (Vekaria & Clack, 1998).

Singh and Singh proposed an enhanced edge recombination crossover for solving the travelling salesman problem (Singh & Singh, 2014). (Larrañaga, et al., 1999) and (Potvin, 1996) presented a review of how to represent the travelling salesman problem, in addition to explaining the advantages and disadvantages of different crossover and mutation operations.

Ray et al. proposed a new algorithm called SWAP_GATSP with two new operators: Knowledge Based Multiple Inversion operator, which was done before selection and Knowledge Based Swapping operator where the process was carried out after the process of crossover. This algorithm was applied to the travelling salesman problem for various data sets, and their results showed superiority as compared to other methods (Ray, et al., 2004).

Kaya and Uyar proposed a new crossover called ring crossover (RC). In this type parents were arranged in the form of a ring, and then a cut point was chosen at random. The other point was the length of the chromosome, the first offspring arises from the point (first cut) in the clockwise direction and the other offspring arises in the counter clockwise direction. They applied this type of crossover to six functions, and obtained results with better performance than the other types of crossover studied (Kaya & Uyar, 2011).

Hong et al, proposed an algorithm called the Dynamic Genetic Algorithm (DGA) in order to apply more than one crossover and mutation at the same time. The algorithm automatically selects the appropriate crossover and appropriate mutation, and also adjusts the crossover and mutation ratios automatically based on the evaluation results of the respective children in the next generation. The algorithm was compared with the simple GA that commonly uses only one crossover process and one process of mutation. Their results showed enhancement of the GA performance (Hong, et al., 2002).

Ahmed, proposed sequential constructive crossover (SCX) to solve the TSP problem, and the basic idea of this method is to choose a point randomly called the crossover site, then employ a method of sequential constructive crossover before the crossover point by using better edges. The rest of the genes after the crossover site are exchanged between parents to get two children; if there is already a gene, then replace it with an unvisited gene (Ahmed, 2010).

Hilding and Ward proposed Automated Operator Selection (AOS) technique, by which they eliminated the difficulties that appear when choosing crossover or mutation operators for any problem. In this work they allowed the genetic algorithm to use more than one crossover and mutation operators, and took advantage of the most effective operators to solve problems. The operators were automatically chosen based on their performance, and therefore the time spent choosing the most suitable operator was reduced. The experiments were performed on the 01-Knapsack problem. This approach proved its effectiveness as compared with the traditional genetic algorithm (Hilding & Ward, 2005).

Dong and Wu proposed a dynamic crossover rate, where the crossover rate is calculated through the ratio between the Euclidean distance between two individuals and the Euclidean distance between the largest and lowest fitness of the individual in the population. The process of crossover between individual "long distance individuals" is thus effective because of differences among themselves, and this would avoid inbreeding and thus overcome premature convergence (Dong & Wu, 2009).

## 3. Proposed Work

The search for the best solution (in genetic algorithms) depends mainly on the creation of new individuals from the old ones. The process of crossover ensures the exchange of genetic material between parents and thus creates chromosomes that are more likely to be better than the parents. There is a large number of crossover methods in the literature, and so the question is: what is the best method to use?

To answer this question, and in an attempt to provide diversity in the population, we have proposed and implemented three types of crossovers, to be compared with the well-known modified crossover.

### 3.1 Cut on worst gene crossover (COWGC)

This method aims to exchange genes between parents by cutting the chromosome at the maximum point that increases the cost. This point (the worst gene) is chosen in both parents depending on the definition of the worst gene for each problem; the worst gene is the point that contributes the maximum to increase the cost of the fitness function of a specific chromosome. For example, the worst gene in the TSP problem is the city with the maximum distance from its left neighbour, while the worst gene in the Knapsack problem is the point with the lowest value to weight ratio, and so on, the "worst" gene is defined based on the problem in hand.

Our algorithm needs to search along the chromosome to find this gene in both parents. The two worst genes are compared to get the worst of both; the index of this point is considered as a cut point in the parent that has the worst gene. The genes after this cut point of the two parents are swapped as in the "Modified Crossover" (Davis, 1985).

Figure1 shows an example of (COWGC). The sign ">" means the worst, i.e. "greater than" if the problem is a minimization problem, and "less than" otherwise.

The cut point (CP) can be calculated for the minimization problem using:

$$CP = \underset{1 \leq i < n}{\operatorname{argmax}}(Distance(C[i], C[i+1])) \quad (1)$$

and for the maximization problem:

$$CP = \underset{1 \leq i < n}{\operatorname{argmin}}(Distance(C[i], C[i+1])) \quad (2)$$

where $C$ represents the chromosome, $i$ is the index of a gene within a chromosome, and the distance function for the TSP can be calculated using either Euclidian distance or using the distances table between cities. The previous equations are used for both parents, and the cut point of the parent that exhibits the maximum distance is used for the crossover operation.

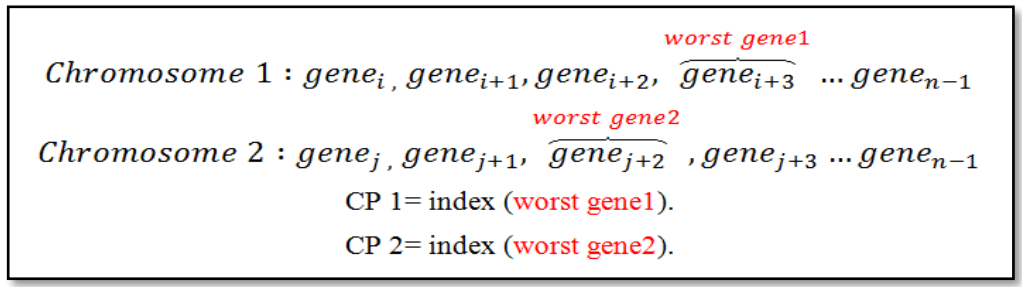

Figure 1. Example of COWGC

**Example 1. (TSP problem)**

If we assume that we represent the TSP problem using path representation and this is natural representation of a tour, each tour is represented by a list of n cities such as: the tour 1-5-4-8-6-7-9 is represented as (1548679).

Figure2 shows examples of the distances of cities. Suppose we choose randomly two parents from the population:

Parent 1: 1 3 8 7 5 6 2 9 4

Parent 2: 1 5 9 8 4 3 7 6 2

| City | 1 | 2 | 3 | 4 | 5 | 6 | 7 | 8 | 9 |
|---|---|---|---|---|---|---|---|---|---|
| 1 | 0 | 2 | 8 | 5 | 20 | 6 | 25 | 30 | 4 |
| 2 |   | 0 | 5 | 3 | 15 | 8 | 52 | 21 | 12 |
| 3 |   |   | 0 | 27 | 6 | 10 | 20 | 14 | 7 |
| 4 |   |   |   | 0 | 8 | 4 | 17 | 60 | 2 |
| 5 |   |   |   |   | 0 | 22 | 6 | 8 | 5 |
| 6 |   |   |   |   |   | 0 | 15 | 6 | 8 |
| 7 |   |   |   |   |   |   | 0 | 10 | 9 |
| 8 |   |   |   |   |   |   |   | 0 | 30 |
| 9 |   |   |   |   |   |   |   |   | 0 |

Figure 2. TSP example distances between cities

To apply COWGC:

- Step1: find the worst gene in the first parent, which is 6, because the distance from 5 to 6 is the maximum and equal to 22 (distance 1), and the worst gene in the second parent is 4, because the distance from 8 to 4 is the maximum and equal to 60 (distance 2).
- Step 2: using equation (1) the cut point of parent 1 is the index (6) and the cut point of parent 2 is index (4).
- Step 3: If (distance1) > (distance2) then
- Apply the Modified crossover in both parents at index (6).
  Else apply the Modified crossover in both parents at index (4) (see Figure3).

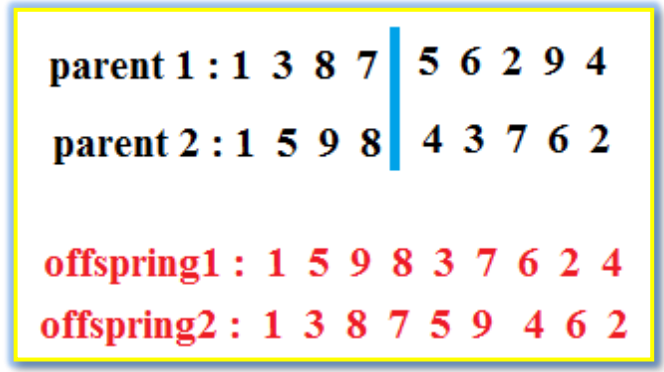

Figure 3. Two offspring output using COWGC

### 3. 2 Cut On Worst L+R Crossover (COWLRGC)

This method is similar to the COWGC, the only difference being that the worst gene is found by calculating the distance between both its neighbours: the right and the left. The cut point can be calculated using:

$$CP = \underset{2 \leq i < n-1}{\mathrm{argmax}}(Distance(C[i], C[i-1]) + Distance(C[i], C[i+1])) \qquad (3)$$

The worst gene is the one that is the sum of the distances with its left and right neighbours and is the maximum among all genes within a chromosome.

**Example 2. (TSP problem)**

Based on Example (1), suppose we choose randomly two parents from the population:

Parent 1: 1 4 2 8 9 6 3 7 5

Parent 2: 1 9 5 7 8 2 3 4 6

To apply COWLRGC:

- Step 1: using equation (3) calculate CP1 for the first parent and CP2 for the second parent; CP1 will be at city "8", because the total distance from city 8 to city 2 and from city 8 to city 9 is the maximum and is = 51 (distance 1). For the second parent, CP2 will be at city 3, because the total distance from city 3 to city 2 and from city 3 to city 4 is the maximum distance and is = 32 (distance 2).
- Step 2: If distance 1 > distance 2 then apply Modified crossover for both parents based on CP1 (city "8"), to create two offspring (see Figure4). Else apply Modified crossover for both parent based on CP2.

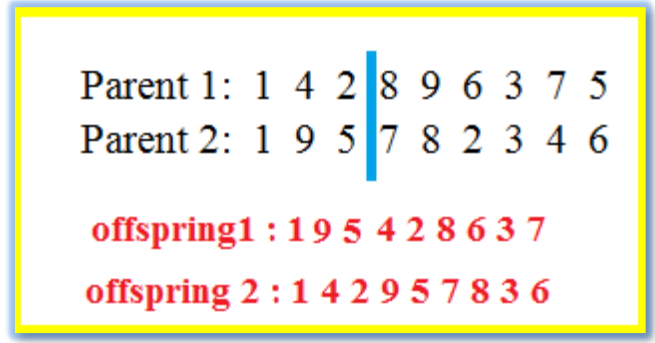

Figure 4. Two offspring output using COWLRGC

### 3.3 Collision Crossover

This type of crossover is inspired by the principle of the head-on elastic collision, when two objects are moving towards each other, with specific velocity and mass for each. If the collision happens, the direction and speed of both objects can be calculated using:

$$v'_1 = \frac{m_1-m_2}{m_1+m_2} v_1 + \frac{2m_2}{m_1+m_2} v_2 \qquad (4)$$

$$v'_2 = \frac{2m_1}{m_1+m_2} v_1 - \frac{m_1-m_2}{m_1+m_2} v_2 \qquad (5)$$

where $v_1$ and $v_2$ are the velocities of objects 1 and 2 respectively, m1 and m2 are the masses of objects 1 and 2 respectively, $v_1$' and $v_2$' are the new velocities after collision of objects 1 and 2 respectively. We assume both objects are moving in the opposite direction, so that one of the velocities should be negative (see Figure5).

Depending on the masses and velocities of the moving objects, the direction and the new velocities can be determined. There are several possibilities after collision, such as: if the 1st object was heavier and faster it will continue in the same direction and so the other object; if both objects are similar they might reflect to the opposite direction or become stationary.

In applying this physics to do crossover, we assume that each gene has its own mass, e.g. masses for Parent 1 = {$m_{11}$, $m_{12}$… $m_{1n}$}, and masses for Parent 2 = {$m_{21}$, $m_{22}$… $m_{2n}$}. Choosing which of these Masses depends on the problem itself; for the TSP, we assumed that each city has a mass, which is its distance from its left and right neighbours. For the 01-knapsack problem, the mass of a gene might be the ratio of its value to its weight.

To do crossover we assume that both chromosomes (parents) are moving towards each other (opposite direction, 180 degrees head-on elastic collision). The velocity of each parent is its total cost, thus each gene within a chromosome has a mass and a velocity.

When both parents collide, each gene (depending on its velocity and mass) will be either reflected, become stationary, or keep moving in the same direction. This can be known from the sign and value of the new velocity.

If the gene is reflected or becomes stationary (v' = 0) this means that the gene is "good", i.e. it is a small distance from its neighbours, and therefore it should remain in its place to form offspring (1). Other genes which carried on moving in the same direction are removed from their places in the new offspring (1), leaving gaps that need to be filled from parent 2. The same procedure is applied to offspring (2).

Equations 4 and 5 decide which remains and which leaves, and the gap places are filled by the other parent, ensuring that no gene is repeated. To foster randomness and diversity we opted for changing the velocity of the moving chromosomes using a random number from 1 to the cost of the chromosome, rather than fixing it to be the chromosome's cost.

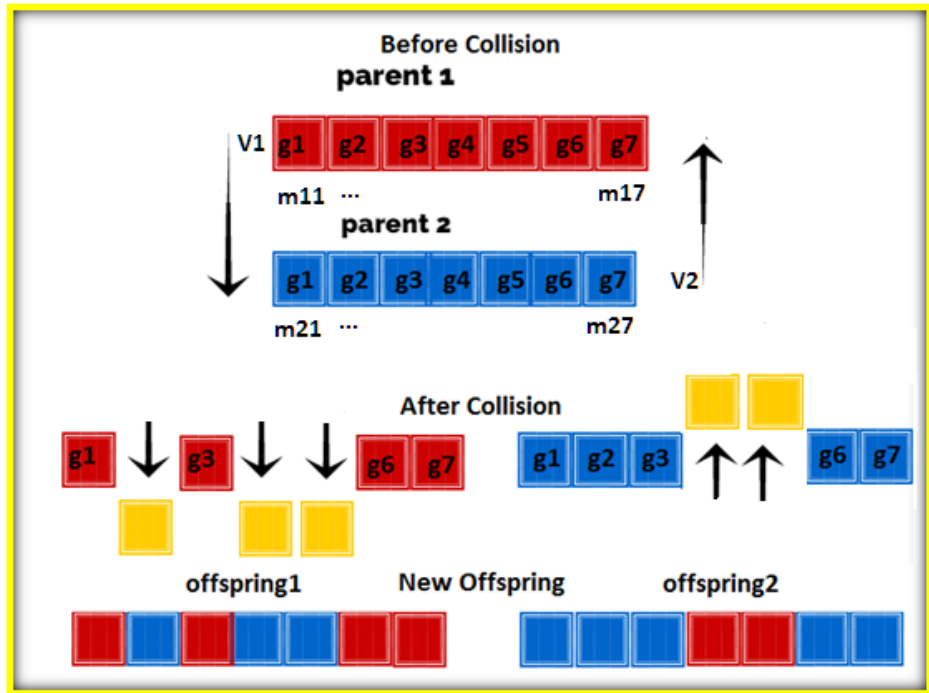

Figure 5. Collision Crossover

### 3.4 Multi Crossover Operator Algorithms

A traditional genetic algorithm commonly uses one crossover operator. We propose using more than one crossover operation, anticipating that different operators will produce different patterns in the offspring and provide some sort of diversity in the population, so as to improve the overall performance of the genetic algorithm.

To do this we opted for two selection approaches: the best cross over, and a randomly chosen crossover.

### 3.4.1 Select the best crossover (SBC)

This algorithm applies multiple crossover operators at the same time on the same parents, and considers the best two offspring to be added to the population, to prevent duplication; only the best and not found in the population is added.

In this work, we applied all the aforementioned crossovers (COWGC, COWLRGC and Collision Crossovers) to the two randomly chosen parents.   The best two offspring that do not already exist in the population are added, though not necessarily from the same operation. This anticipates that such a process encourages diversity in the population, and thus avoids falling into local optima.

### 3.4.2 Select any crossover (SAC)

This algorithm is similar to SBC, the difference lying in applying only one crossover operator each time; the selection strategy is random.

Randomly choose one of the aforementioned crossovers (COWGC, COWLRGC and Collision Crossovers) in the two randomly chosen parents, and add the two new offspring to the population. We reckon that in each generation or so, the algorithm chooses a different type of crossover. This means that different types of offspring will result, and this is what we are aiming for, thus increasing diversification in the population, and attempting to enhance the performance of the genetic algorithm.

## 4. Results and Discussion

To evaluate the proposed methods, we conducted two sets of experiment on different TSP problems; both were designed to compare the performance of the proposed crossover operators with the well-known Modified crossover (Davis, 1985) and PMX crossover (Goldberg & Lingle, 1985), using simulated and real data.

### 4.1 First set of experiments

The aim of those experiments is to measure the effectiveness of the proposed crossover operators (COWGC, COWLRGC and Collision Crossovers) in terms of convergence time. The results of those experiments were compared to the well-known crossover (the one-point modified crossover) and PMX crossover, in addition to measuring the performance of the genetic algorithm when using either SAC or SBC.

These crossover operators were tested using three test data: the first was random cities, where the coordinates of the cities were chosen randomly, and the second and third test data are real data – "bier127" and "a280" taken from TSPLIB (Reinelt & Gerhard, 1996), each of them consisting of 100, 127, and 280 cities respectively. Mutation used in this experiment is the Exchange mutation (Banzhaf, 1990).

The genetic algorithm parameters that were selected in the first test included the following: population size: 100, the probability of crossover is (83%), mutation probability is (2%), and the maximum generation was 2000. The results of the first test are shown in Figures (6, 7, and 8).

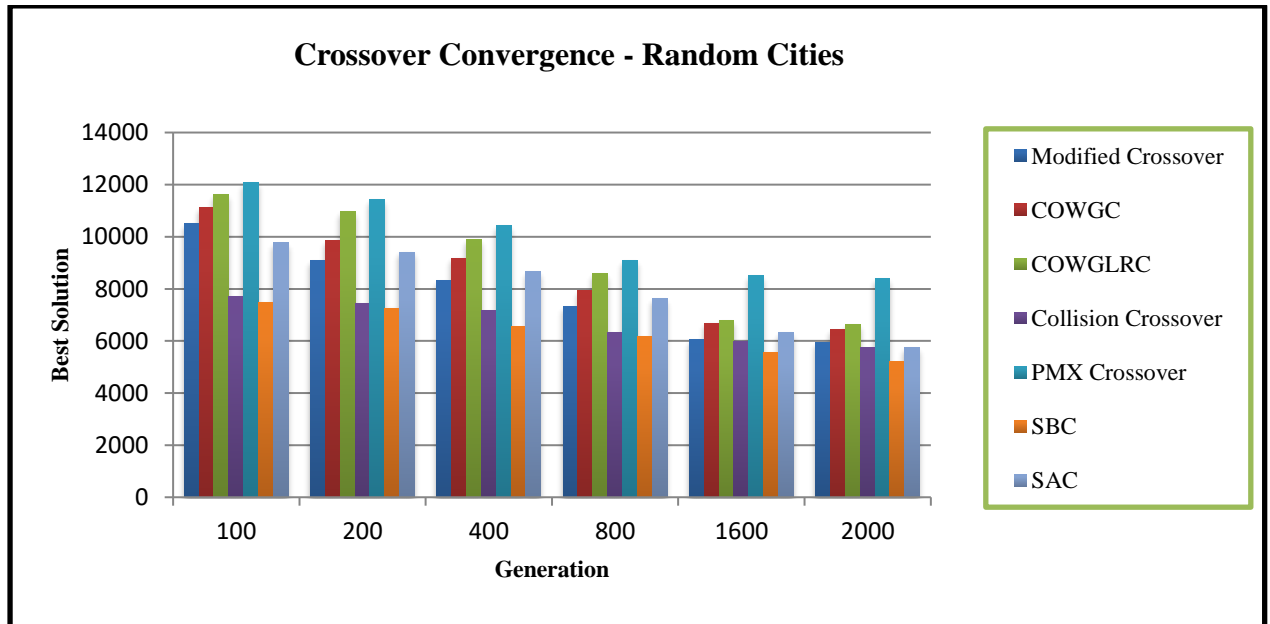

**Figure 6. Crossover convergence with crossover ratio = 0.83 (random cities)**

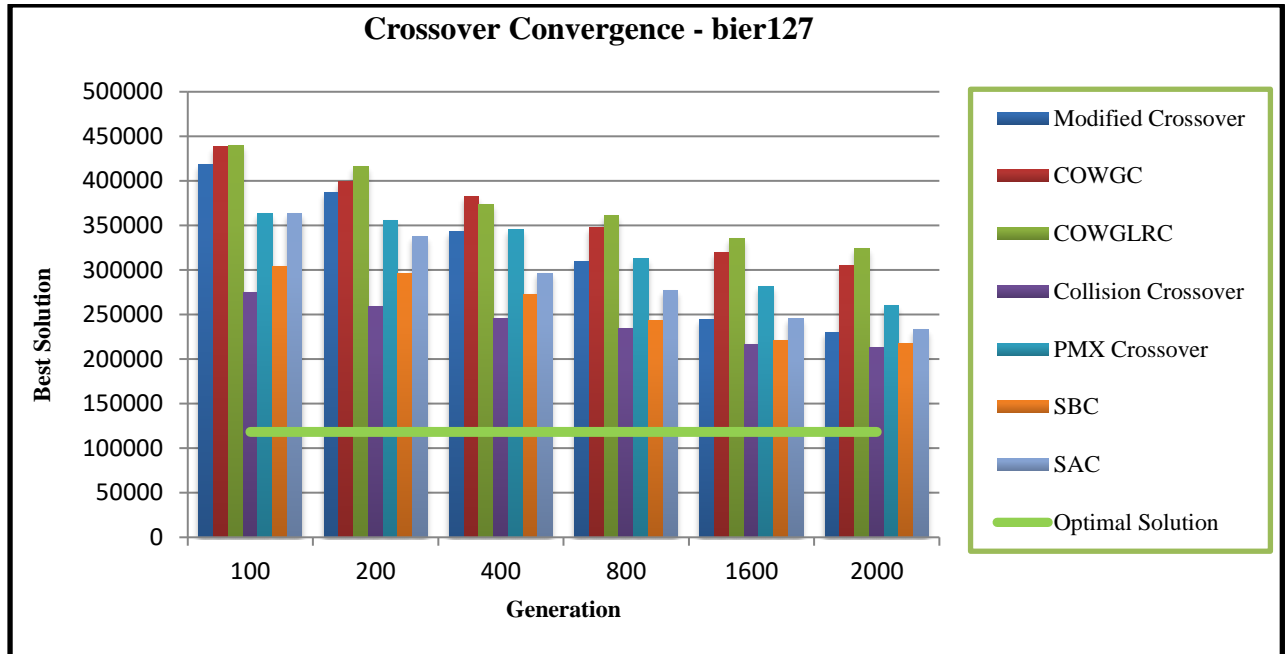

Figure 7. Crossover convergence with crossover ratio = 0.83 (bier127)

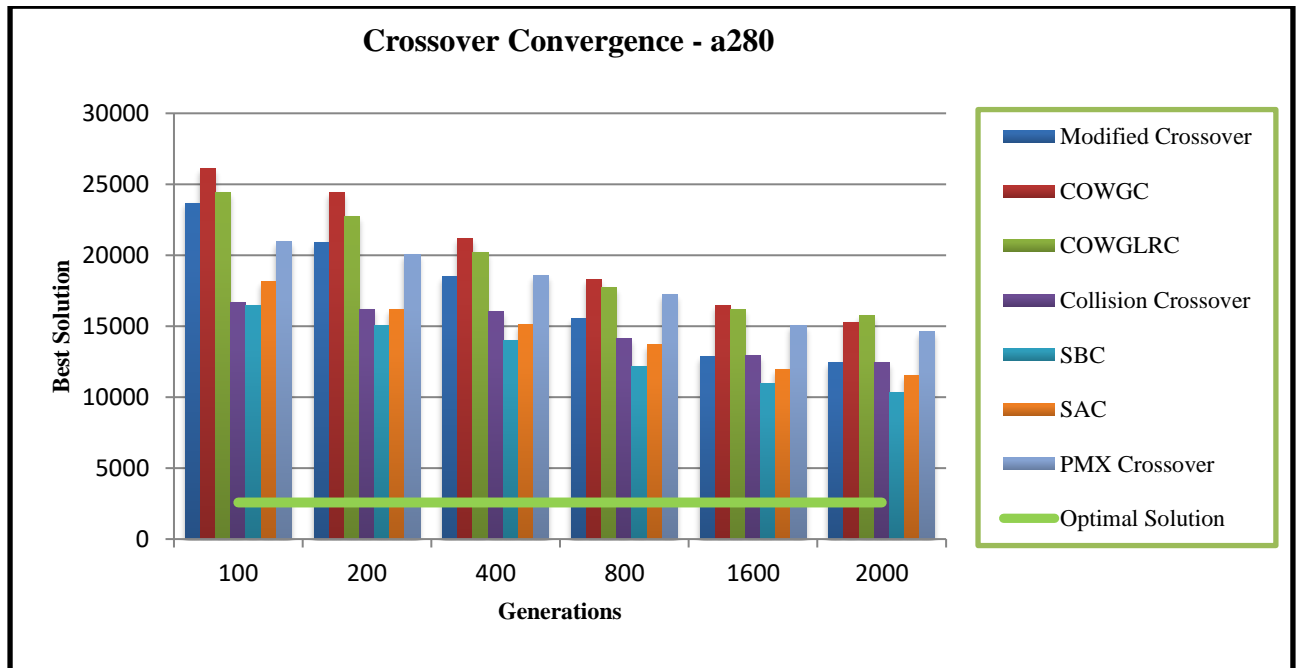

Figure 8. Crossover convergence with crossover ratio = 0.83 (a280)

In the second test we used the same parameters, applied to the same problems, but the crossover probability was increased to (92%). The results of the second test are shown in Figures (9, 10, and 11). In this test we also calculated the consumed time of each of the five operators, and selection algorithm. The execution time result is shown in Table (1).

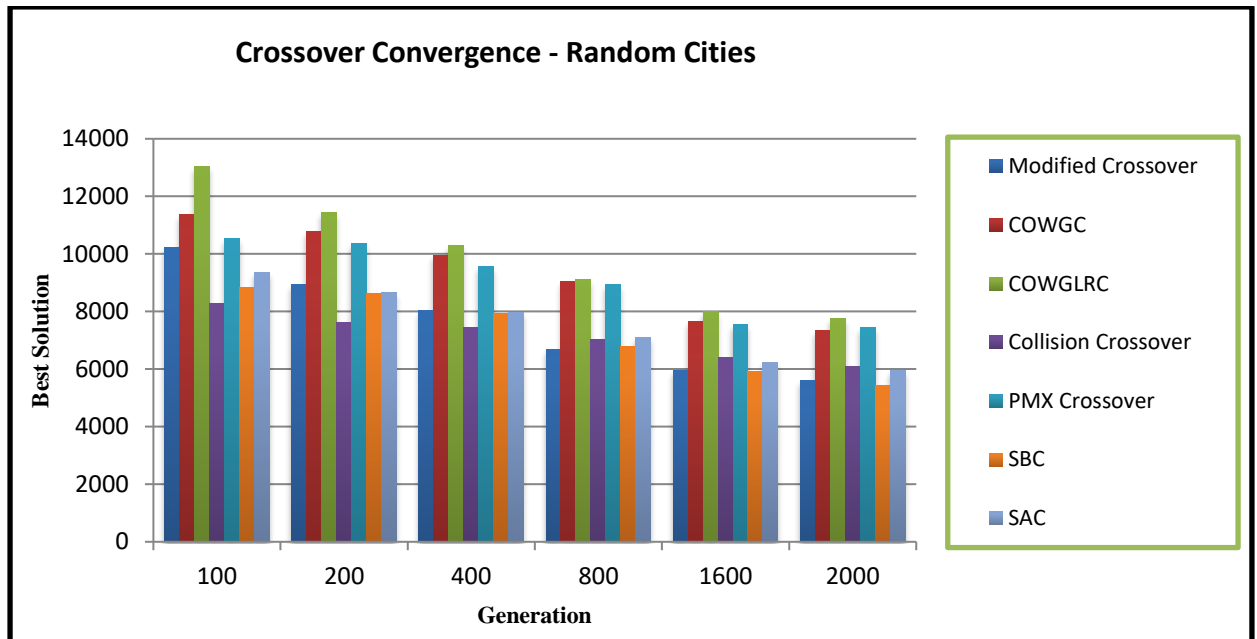

Figure 9. Crossover convergence with crossover ratio = 0.92 (random cities)

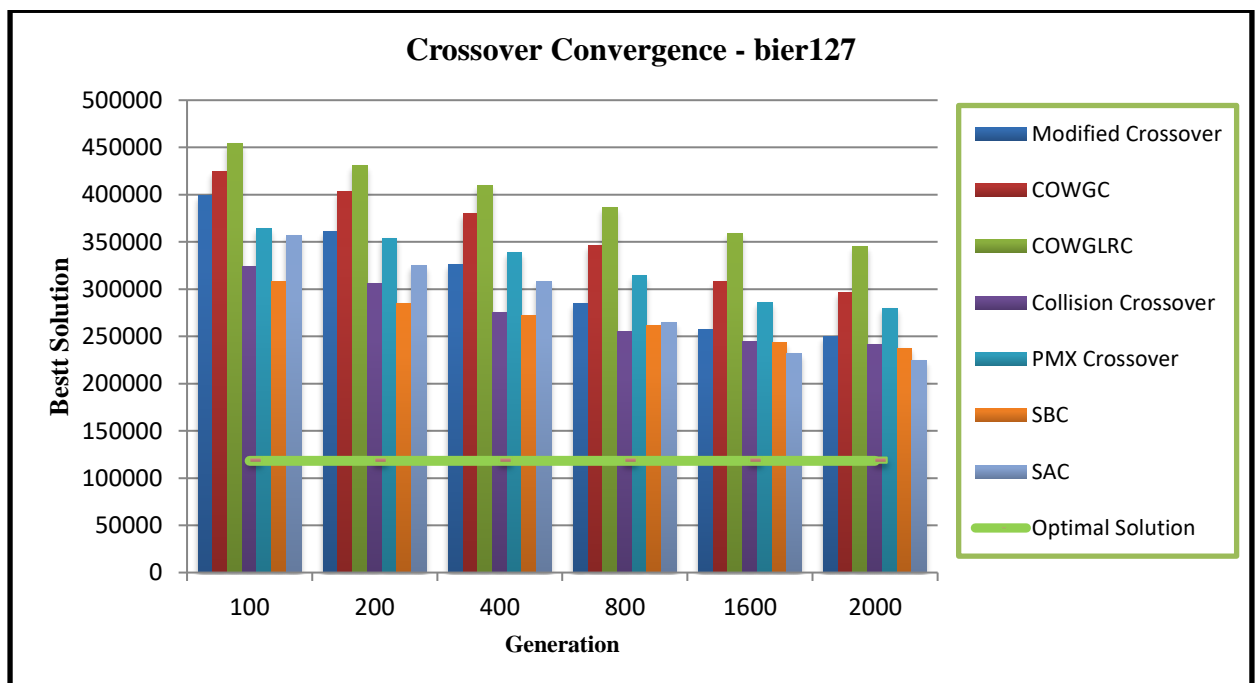

Figure 10. Crossover convergence with crossover ratio = 0.92 (bier127)

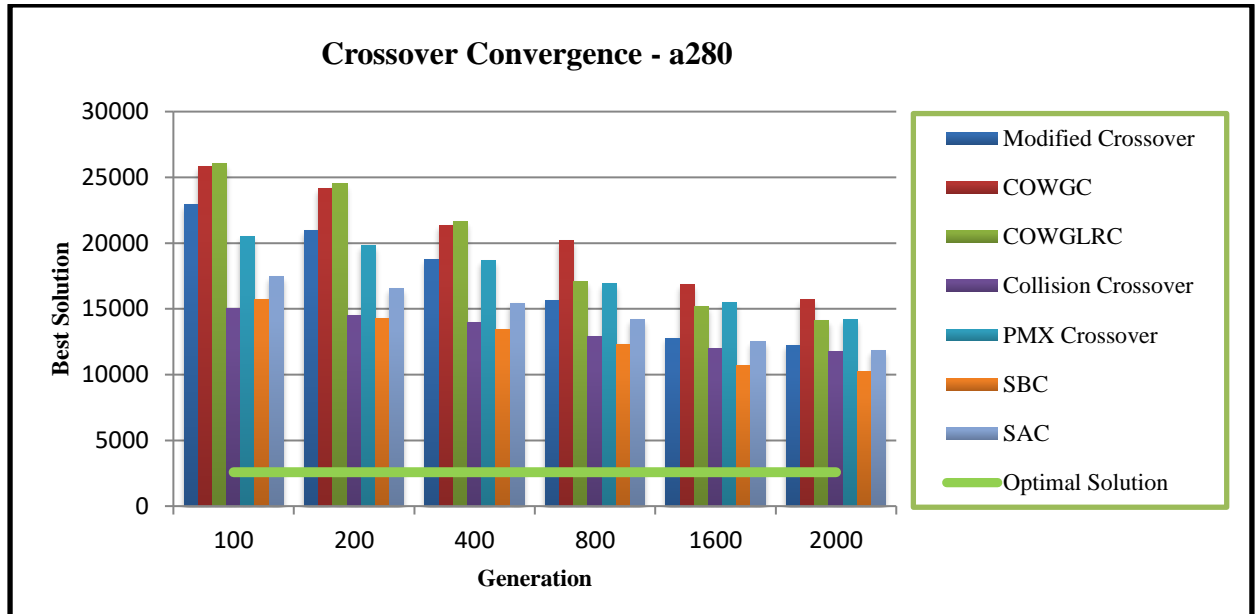

Figure 11. Crossover convergence with crossover ratio = 0.92 (a280)

Results from the first test indicate that the best performance is recorded by the SBC, followed by the Collision crossover and SAC, in most cases the performance of the Collision crossover was better than that of the SAC, because it showed good convergence to a minimum value. The Collision crossover showed a faster convergence rate than other crossover operators, followed by the modified crossover. The performance of each method is shown in Figures (6), (7) and (8).

Increasing the crossover ratio (the second test) does not enhance the performance of all the method significantly. A closer look at Figures (6-11) reveals that the SBC algorithm outperformed all other methods in the speed of convergence. In addition, the Collision crossover and the modified crossover showed rapid convergence to a near optimal solution as compared to other methods.

Despite the SBC outperforming the Collision crossover and SAC, the Collision crossover and SAC is still better than the SBC in terms of the time consumed (see Table (1)), because SBC tries all the available crossovers and selects the best, which consumes more time. While the SAC selects any one randomly, and the Collision crossover is only one crossover operator by definition. Moreover, the differences between the results of the three methods are not significant. In addition, if the number of crossover operators in the SBC and SAC increased, both algorithms might become more efficient, but this is at the expense of the SBC's consumed time.

Table 1. The execution time for each method.

| Problem | Modified Crossover | COWGC | COWGLRC | Collision Crossover | PMX Crossover | SBC | SAC |
|---|---|---|---|---|---|---|---|
| rat783 | **174396** | 251066 | 241951 | 208201 | 783933 | 196179 | 216304 |
| a280 | **80910** | 88865 | 112087 | 101781 | 278297 | 87003 | 80902 |
| u159 | 61385 | 72646 | 63307 | **55242** | 169495 | 72108 | 58410 |
| ch130 | **46764** | 57155 | 67489 | 58262 | 148853 | 65615 | 55791 |
| bier127 | **52898** | 62677 | 72838 | 59921 | 153931 | 59937 | 54600 |
| kroA100 | 45022 | **43618** | 51505 | 52342 | 137712 | 55969 | 55225 |
| pr76 | **44653** | 49864 | 56936 | 49056 | 114433 | 45887 | 49411 |
| berlin52 | **33303** | 43620 | 41760 | 40801 | 68828 | 38786 | 42816 |
| att48 | 35569 | 34583 | 38366 | 46029 | 65566 | 41750 | **31623** |

| eil51 | 36687 | 45380 | 46540 | 37240 | 83595 | 36109 | **32214** |
| pr144 | 57671 | 76635 | 69275 | 68074 | 165854 | 68419 | **53571** |

Some crossovers showed better performance than others, and this does not mean that the rest of the crossovers have proved their failure, as they can be effective when used by SBC and SAC, where they urge diversity through the different patterns of individuals and hence increases the efficiency of both algorithms and help escaping local minima.

### 4.2 Second set of experiments

The aim of those experiments is to measure the effectiveness of the winners (SBC, Collision crossover and SAC operators) compared to the optimal solutions. These algorithms, in addition to the Modified crossover and PMX crossover, were tested using eleven TSP problems taken from the TSPLIB, those include rat783, a280, u159, ch130 bier127, kroA100, pr76, berlin52, att48, eil51 and pr144 (the numbers attached to the problem names represent the number of cities).

The genetic algorithm parameters, which were selected for the first test for all algorithms include the following: the crossover ratio is 100%, mutation ratio is 0%, the size of population is 200, and the maximum number of generations is 8000 (see Table (2)).

**Table 2. Comparison of crossover methods on converging to optimal solution (population size = 200)**

| Problem | SBC | SAC | Collision crossover | PMX crossover | Modified crossover | Optimal Solution |
|---|---|---|---|---|---|---|
| rat783 | 99581 | 104474 | **96647** | 143981 | 146769 | **8806** |
| a280 | **16504** | 16881 | 17048 | 25131 | 24172 | **2579** |
| u159 | 180580 | 213555 | **168037** | 291877 | 292269 | **42080** |
| ch130 | 22407 | 24652 | **18724** | 29692 | 31834 | **6110** |
| bier127 | **313554** | 335392 | 332013 | 433551 | 420988 | **118282** |
| kroA100 | **53572** | 74023 | 57515 | 97490 | 97013 | **21282** |
| pr76 | 226251 | 248414 | **214962** | 331358 | 320816 | **108159** |
| berlin52 | 11409 | 13924 | **10224** | 16178 | 16142 | **7542** |
| att48 | **47456** | 70669 | 51439 | 72269 | 75278 | **10628** |
| eil51 | **642** | 735 | 649 | 949 | 878 | **426** |
| pr144 | **309973** | 379911 | 335088 | 480611 | 504435 | **58537** |
| Average | 116539 | 134784.5 | 118395.1 | 174826.1 | 175508.5 | - |

In the second test, the same previous parameters were used except for the size of population, which was reduced to 100 (see Table (3)).

As can be seen from Table (2), the results indicate the superiority of the SBC and the Collision crossover over the other methods, where the SBC performed the best for the problems: a280, bier127, kroA100, att48, eil51 and pr144. And the Collision crossover performed the best for the rest of the problems.

**Table 3. Comparison of crossover methods on converging to optimal solution (population size = 100)**

| Problem | SBC | SAC | Collision crossover | PMX crossover | Modified crossover | Optimal Solution |
|---|---|---|---|---|---|---|
| rat783 | 99076 | 100301 | **96895** | 149925 | 152036 | **8806** |
| a280 | **15157** | 16928 | 15335 | 25554 | 23027 | **2579** |
| u159 | 196521 | **190217** | 192696 | 304470 | 277389 | **42080** |
| ch130 | 20423 | 20942 | **20310** | 29501 | 29956 | **6110** |
| bier127 | 333549 | 348203 | **324922** | 456519 | 412957 | **118282** |
| kroA100 | **62968** | 81001 | 63269 | 98378 | 95493 | **21282** |
| pr76 | **239327** | 254548 | 240782 | 336533 | 340641 | **108159** |
| berlin52 | 13240 | 12336 | **11798** | 16456 | 14965 | **7542** |
| **att48** | 51959 | 60126 | **46315** | 77061 | 67630 | **10628** |
| **eil51** | 673 | 753 | **556** | 918 | 827 | **426** |
| **pr144** | 350733 | 333256 | **279038** | 455153 | 539113 | **58537** |
| **Average** | **125784.2** | **128964.6** | **117446.9** | **177315.3** | **177639.5** | **-** |

By reducing the population size to 100, the results degraded a bit as it can be seen from Table (3). Interestingly, the best performance was recorded by the Collision crossover, which outperformed the other methods in seven problems: rat783, ch130, bier127, berlin52, att48, eil51, pr144 followed by the SBC algorithm and SAC algorithm. The Modified crossover and PMX method showed slow convergence to near optimal solutions.

Having known that the Collision crossover came the second in the first set of experiments and the first in the second set of experiments in terms of finding the minimum solution and the fastest convergence (best solution in less number of generations), this put the proposed Collision crossover at the top of the compared methods. The good performance of the Collision crossover is due to the randomness of the collided genes that output offspring that are significantly different from their parents, and this serves as an alternative to the mutation operator, which is used to escape local minima.

Some of the solutions produced by the tested algorithms were close to the optimal solutions, but none could achieve an optimal solution. This shows the importance of using appropriate parameters along with crossover (such as population size, a higher mutation ratio) and appropriate number of generations, due to the effective impact of their convergence to the optimal solution.

## 5. Conclusions

We have proposed several crossover methods (COWGC, COWLRGC and Collision Crossovers), in addition to proposing selection approaches (SBC and SAC). Several experiments were conducted to evaluate those methods on several TSP problems, which showed the efficiency of some of the proposed methods (namely, Collision Crossovers, SBC and SAC) over the well-known Modified crossover method and PMX crossover. The proposed Collision Crossovers performed the best in terms of accuracy and time consuming, and this is due to the randomness of the collided genes that output offspring that are significantly different from their parents.

This study also shows that using more than one crossover method in GA can enhance its performance, because it allows the GA to avoid the local minima; the proposed SBC and SAC strategies enhance the performance of GA. This is due to the diversity of the solutions in the population maintained by different crossover operators employed by both of the SBC and SAC. However, taking performance, convergence speed and consumed time into consideration, we found that the proposed Collision crossover (in general) is the best choice for GA.

Crossover alone is not enough to enhance the performance of GA, therefore our future work will include the development of some types of new Mutations, using the same approaches, i.e. trying more than one mutation each time to support the proposed approaches and attempting to further enhance the performance of GA (Hassanat, et al., 2016). Additionally, we will apply the proposed methods (particularly the collision crossover) to different problems using different benchmark data, such as the Knapsack problem and other optimization problems.

### Acknowledgements

The first author made part of his contribution to the work presented in this paper while he was a Visiting Research Fellow at the Sarajevo School of Science and Technology (www.ssst.edu.ba). He thanks the University for hosting his visit and for all the support and assistance afforded to him during the visit.

### References

Ahmed, Z. H., 2010. Genetic algorithm for the traveling salesman problem using sequential constructive crossover operator. *International Journal of Biometrics & Bioinformatics (IJBB),* 3(6), p. 96.

Amsaveni, A. & Vanathi, P. T., 2015. An efficient reversible data hiding approach for colour images based on Gaussian weighted prediction error expansion and genetic algorithm. *International Journal of Advanced Intelligence Paradigms,* 7(2), pp. 156-171.

Ayala, H. V. H. & dos Santos Coelho, L., 2012. Tuning of PID controller based on a multiobjective genetic algorithm applied to a robotic manipulator. *Expert Systems with Applications,* 39(10), pp. 8968-8974.

Bäck, T. & Schwefel, H. P., 1993. An overview of evolutionary algorithms for parameter optimization. *Evolutionary computation,* 1(1), pp. 1-23.

Banzhaf, W., 1990. The "molecular" traveling salesman. *Biological Cybernetics,* 64(1), pp. 7-14.

Benkhellat, Z. & Belmehdi, A., 2012. *Genetic algorithms in speech recognition systems.* s.l., s.n., pp. 853-858.

Chen, Y., Liu, Y., Wang, C. & Ma, R., 2015. Associations between population topologies and Gaussian dynamic particle swarm performance. *International Journal of Modelling, Identification and Control,* 24(2), pp. 138-148.

Davis, L., 1985. Applying adaptive algorithms to epistatic domains. *IJCAI,* August, Volume 85, pp. 162-164.

Deb, K. & Agrawal, S., 1999. Understanding interactions among genetic algorithm parameters. *Foundations of Genetic Algorithms,* pp. 265-286.

Deep, K. & Thakur, M., 2007. A new crossover operator for real coded genetic algorithms. *Applied mathematics and computation,* 188(1), pp. 895-911.

Dong, M. & Wu, Y., 2009. Dynamic Crossover and Mutation Genetic Algorithm Based on Expansion Sampling. *Artificial Intelligence and Computational Intelligence,* pp. 141-149.

Eiben, A. E., Michalewicz, Z., Schoenauer, M. & Smith, J. E., 2007. Parameter control in evolutionary algorithms. *Parameter setting in evolutionary algorithms,* pp. 19-46.

Eiben, A. E. & Smith, J. E., 2003. *Introduction to evolutionary computing.* s.l.:Springer Science & Business Media.

Gallard, R. H. & Esquivel, S. C., 2001. Enhancing evolutionary algorithms through recombination and parallelism. *Journal of Computer Science & Technology,* Volume 1.

Goldberg, D. E., 1989. *Genetic algorithms in search, optimization, and machine learning.* s.l.:Addison-wesley Reading Menlo Park.

Goldberg, D. E. & Lingle, R., 1985. s.l., Lawrence Erlbaum Associates, pp. 154-159.

Grefenstette, J., Gopal, R., Rosmaita, B. & Van Gucht, D., 1985. *Genetic algorithms for the traveling salesman problem.* s.l., s.n., pp. 160-168.

Gupta, H. & Wadhwa, D. S., 2014. Speech feature extraction and recognition using genetic algorithm. *International Journal of Emerging,* 4(1).

Hassanat, A. et al., 2016. Enhancing Genetic Algorithms using Multi Mutations. *arXiv 1602.08313.*

Hilding, F. & Ward, K., 2005. *Automated Crossover and Mutation Operator Selection on Genetic Algorithms.* Melbourne, s.n., pp. 903-909.

Holland, J. H., 1975. *Adaptation in natural and artificial systems: an introductory analysis with applications to biology, control, and artificial intelligence.* Cambridge, MA: MIT Press.

Hong, T. P., Wang, H. S., Lin, W. Y. & Lee, W. Y., 2002. Evolution of appropriate crossover and mutation operators in a genetic process. *Applied Intelligence,* 16(1), pp. 7-17.

Kantha, A., Utkarsh, A. & Jatoth, R. K., 2016. Hybrid genetic algorithm-swarm intelligence-based tuning of temperature controller for continuously stirred tank reactor. *International Journal of Modelling, Identification and Control,* 25(3), pp. 239--248.

Kaya, Y. & Uyar, M., 2011. A novel crossover operator for genetic algorithms: ring crossover. *arXiv preprint arXiv:1105.0355.*

Kotenko, I. & Saenko, I., 2015. Improved genetic algorithms for solving the optimisation tasks for design of access control schemes in computer networks. *International Journal of Bio-Inspired Computation,* 7(2), pp. 98-110.

Larrañaga, P. et al., 1999. Genetic algorithms for the travelling salesman problem: A review of representations and operators. *Artificial Intelligence Review,* 13(2), pp. 129-170.

Mahmoudi, A. & Mahmoudi, S., 2014. Solving Assignment Problem Using Genetic and MMAS Algorithms. *Caspian Journal of Applied Sciences Research,* 3(8), pp. 1-8.

Man, K. F., Tang, K. S. & Kwong, S., 1996. Genetic Algorithms: Concepts and Applications. *IEEE Transactions on Industrial Electronics,* 43(5), pp. 519-534.

Mustafa, W., 2003. Optimization of Production Systems Using Genetic Algorithms. *International Journal of Computational Intelligence and Applications,* 3(03), pp. 233-248.

Nakamura, H., 1997. Optimization of Facility Planning and Circuit Routing for Survivable Transport Networks─An Approach Based on Genetic Algorithm and Incremental Assignment. *IEICE Transactions on Communications,* 80(2), pp. 240-251.

Oliver, I. M., Smith, D. & Holland, J. R., 1987. *Study of permutation crossover operators on the traveling salesman problem.* s.l., s.n.

Paulinas, M. & Ušinskas, A., 2015. A survey of genetic algorithms applications for image enhancement and segmentation. *Information Technology and control,* 36(3).

Potvin, J. Y., 1996. Genetic algorithms for the traveling salesman problem. *Annals of Operations Research,* 63(3), pp. 337-370.

Ray, S. S., Bandyopadhyay, S. & Pal, S. K., 2004. *New operators of genetic algorithms for traveling salesman problem.* s.l., IEEE, pp. 497-500.

Reinelt & Gerhard, 1996. *TSPLIB. University of Heidelberg.* [Online] Available at: http://comopt.ifi.uni-heidelberg.de/software/TSPLIB95 [Accessed 17 9 2015].

Singh, A. & Singh, R., 2014. Exploring Travelling Salesman Problem using Genetic Algorithm. *International Journal of Engineering Research & Technology (IJERT),* 3(2).

Spears, W. M., 1992. Crossover or mutation. *Foundations of genetic algorithms,* Volume 2, pp. 221-237.

Spears, W. M., 1995. *Adapting Crossover in Evolutionary Algorithms.* s.l., s.n., pp. 367-384.

Spears, W. M. & De Jong, K. A., 1990. An analysis of multi-point crossover. *Foundations of genetic algorithms.*

Srivastava, P. R. & Kim, T. H., 2009. Application of genetic algorithm in software testing. *International Journal of software Engineering and its Applications,* 3(4), pp. 87-96.

Syswerda, G., 1989. *Uniform crossover in genetic algorithms.* s.l., s.n., pp. 2-9.

Tian, Z., Li, S. & Wang, Y., 2016. T-S fuzzy neural network predictive control for burning zone temperature in rotary kiln with improved hierarchical genetic algorithm. *International Journal of Modelling, Identification and Control,* 25(4), pp. 323 - 334.

Tsang, P. & Au, A., 1996. *A genetic algorithm for projective invariant object recognition.* s.l., IEEE, pp. 58-63.

Vekaria, K. & Clack, C., 1998. *Selective crossover in genetic algorithms: An empirical study.* s.l., Springer, pp. 438-447.

Zhong, J., Hu, X., Gu, M. & Zhang, J., 2005. *Comparison of performance between different selection strategies on simple genetic algorithms.* s.l., IEEE, pp. 1115-1121.